\documentclass{bmvc2k}

\usepackage{arydshln}
\usepackage[dvipsnames]{xcolor}
\usepackage{multirow}
\usepackage{graphbox,graphicx}
\usepackage{colortbl}
\usepackage{amssymb}
\usepackage[export]{adjustbox}
\usepackage[noline,boxed]{algorithm2e}

\newcommand{\myparagraph}[1]{\noindent \textbf{#1}\ }
\newcommand{\camr}[1]{{\color{black} #1}}

\title{Class Interference Regularization}

\addauthor{Bharti Munjal}{munjalbharti@gmail.com}{1}
\addauthor{Sikandar Amin}{sikandaramin@gmail.com}{1}
\addauthor{Fabio Galasso}{galasso@di.uniroma1.it}{2}

\addinstitution{
 OSRAM GmbH, \\ 
 Parkring 33 Garching,\\
 Munich 85748, Germany
}

\addinstitution{
 Sapienza University of Rome, \\ 
 Piazzale Aldo Moro, 5,\\
 Rome 00185, Italy
}

\runninghead{B.MUNJAL ET. AL}{CLASS INTERFERENCE REGULARIZATION}

\def\eg{\emph{e.g}\bmvaOneDot}

\begin{document}

\maketitle

\begin{abstract}
Contrastive losses yield state-of-the-art performance for person re-identification, face verification and few shot learning. They have recently outperformed the cross-entropy loss on classification at the ImageNet scale and outperformed all self-supervision prior results by a large margin (SimCLR). Simple and effective regularization techniques such as label smoothing and self-distillation do not apply anymore, because they act on multinomial label distributions, adopted in cross-entropy losses, and not on tuple comparative terms, which characterize the contrastive losses.

Here we propose a novel, simple and effective regularization technique, the \emph{Class Interference Regularization} (CIR), which applies to cross-entropy losses but is especially effective on contrastive losses. CIR perturbs the output features by randomly moving them towards the average embeddings of the negative classes. To the best of our knowledge, CIR is the first regularization technique to act on the output features.

In experimental evaluation, the combination of CIR and a plain Siamese-net with triplet loss yields best few-shot learning performance on the challenging tieredImageNet. CIR also improves the state-of-the-art technique in person re-identification on the Market-1501 dataset, based on triplet loss, and the state-of-the-art technique in person search on the CUHK-SYSU dataset, based on a cross-entropy loss. Finally, on the task of classification CIR performs on par with the popular label smoothing, as demonstrated for CIFAR-10 and -100.

\end{abstract}
\vspace{-0.5cm}

\section{Introduction}
\label{sec:intro}

Contrastive losses yield state-of-the-art performance for person re-identification~\cite{DBLP:journals/corr/HermansBL17, Chen_2019_ICCV, Guo_2019_ICCV, zhou2019osnet, Auto2019Ruijie}, face verification~\cite{Schroff_2015_CVPR, chopra2005learning} and few shot learning~\cite{wang2018large}. In their general formulation, contrastive losses imply processing the input samples with Siamese networks, then penalizing them if the output embeddings of two samples from the same class (\emph{aka} positives) have higher distances than those from different classes (\emph{aka} negatives). Most recent advances in self-supervised training for classification~\cite{MoCo,PIRL} have leveraged contrastive losses and this is also the case for the current best method, SimCLR~\cite{SimCLR}, which has set aside from the competition by a large margin at scale on ImageNet~\cite{imagenet_cvpr09}. Notably, at the moment of writing, a novel technique based on contrastive loss~\cite{SupervisedContrL} has just achieved best supervised-learning performance on ImageNet, outperforming for the first time the established cross-entropy loss.

In this work, we propose a novel, simple and effective regularization technique, the \emph{Class Interference Regularization} (CIR), which applies to models trained with the traditional cross-entropy loss, but also and most effectively in the case of contrastive losses such as the triplet loss. \camr{CIR introduces a data-driven noise term}. It works by estimating output features for each sample image in the batch and then randomly perturbing them with the average embeddings of their negative classes.

CIR fills in a gap in the training of neural networks with contrastive losses, because the widely adopted and effective label smoothing~\cite{christian2016cvpr} and self-distillation~\cite{Yang2019TrainingDN} do not apply to those. Both the techniques ease the training with cross-entropy by perturbing the axis-aligned label distributions, also termed ``one-hot vectors''. The first moves the labels off-axis; the second adopts soft-labels from prior rounds of training. Both of the techniques have been studied in-depth and have been widely adopted~\cite{pereyra2017regularizing,rafael2019nips,li2020Regularization,Furlanello2018BornAN}. However none of them applies in the case of contrastive losses because they require multinomial label distributions, while contrastive terms such as triplets use comparative embedding distances between (positive and negative) samples. By contrast, CIR applies both for cross-entropy and contrastive losses.

We thoroughly experiment with CIR in the case of contrastive and cross-entropy losses. Best performance improvements are achieved for the first. In particular, CIR sets a new state-of-the-art performance for few-shot learning on tieredImageNet~\cite{ren18fewshotssl}, with a plain Siamese-net and a triplet loss, reaching 69.1\% 1-shot and 82.9\% 5-shot accuracies, with absolute margins of 2.8pp and 1.4pp on the second best respectively. Also CIR improves the plain Siamese-net + triplet in the case of person re-identification on the Market-1501~\cite{zheng2015scalable} and also slightly improves the best performer ABD-Net~\cite{Chen_2019_ICCV}. In the case of cross-entropy losses, CIR improves slightly but consistently a state-of-the-art person search technique~\cite{munjal2019bmvc} on the CUHK-SYSU dataset~\cite{xiao2017joint} and it yields better classification on the CIFAR-10 and -100 datasets~\cite{cifar10}, on par with the established label smoothing~\cite{christian2016cvpr}.

\section{Related Work}
\label{sec:related}

\textbf{\textit{Regularization}} is a major topic when learning over-parameterized models such as Deep Neural Networks (DNN). We review most relevant, recent and widely-adopted techniques by grouping them intro three broad categories, depending on their application focus.

\noindent \textbf{Input samples.} Most common methods in this category regularize the training by data augmentation, i.e.\ by performing random transformations of the input samples such as cropping, rotation, flipping, noise injection or random erasing~\cite{bishop1995training, journals/jmlr/VincentLLBM10,zhong2020random}. More complex methods use GANs to generate synthetic data~\cite{DBLP:journals/corr/abs-1712-04621} or add adversarial examples~\cite{DBLP:journals/corr/GoodfellowSS14} to the training set.

\noindent \textbf{Network weights and hidden units.}
Most popular techniques are weight decay~\cite{Krizhevsky09learningmultiple} and dropout~\cite{JMLR:v15:srivastava14a}. The first adds $\ell_1$ or $\ell_2$ norms of the weights into the loss, to bias training towards simpler solutions. The second randomly drops neurons to avoid weight co-adaptation, as also targeted by the variants DropConnect~\cite{pmlr-v28-wan13} and Adaptive Dropout~\cite{NIPS2013_5032}.
Other technique regularize via stochastic pooling~\cite{zielerpooling}, depth~\cite{stochDepth}, or by integrating adversarial noise layers in the CNN~\cite{ANL}.

\noindent \textbf{Label distributions.} Most utilized in this category is Label Smoothing~\cite{christian2016cvpr, rafael2019nips} that moves the axis-aligned target label distribution (``one-hot vectors'') off the axis, thus softening it.
Also widely adopted is Self-Distillation~\cite{NIPS2015_distilling, Yang2019TrainingDN} which softens the labels via iterations of trainings on generations of network predictions.
Also in this category are Disturb Label~\cite{xie2016disturblabel}, which randomly flips the ground-truth labels of some input images into wrong ones, and Mixup~\cite{zhang2018mixup}, changing the target label by mixing input images in known proportions.

CIR differs from all of the above because it applies to the output features of a DNN. CIR is closest in spirit to techniques which regularize via label distributions. However established techniques such as label smoothing and self-distillation only apply to cross-entropy losses, while CIR applies to both cross-entropy and contrastive losses. \camr{Also, CIR is data-driven, similarly to \eg data augmentation techniques using PCA on the RGB pixels values~\cite{cifar10}. But CIR is the first to act on the output space.}

\textbf{\textit{Multiple tasks}} are here considered to thoroughly evaluate the benefits of CIR, which we also briefly review for related work.

\noindent\textbf{Few-Shot Learning.}
This task targets classification of query samples from a single or few training (\textit{aka} support) samples. Methods are broadly split into optimization- and metric-based.
The first follow from MAML~\cite{DBLP:journals/corr/FinnAL17} and aim to learn good initial parameters of a learner to adapt with gradient descent~\cite{Ravi2017OptimizationAA,DBLP:journals/corr/abs-1803-02999,DBLP:journals/corr/LiZCL17,mishra2018a,jiang2018learning,rusu2018metalearning}.
The second learn a common embedding space for both the support and query samples~\cite{garcia2019graph,han2019meta,ren2018meta,sung2018cvpr,oriol2016nips,chen2019closerlook, wang2018large} and have been popularized by prototypical networks~\cite{snell2017nips}. We apply CIR to improve the prototypical network with triplet loss technique of \cite{wang2018large} and achieve state-of-the-art results.

\noindent \textbf{Person Re-Identification.}
This targets retrieving a (query) person identity from a gallery of individuals, using cropped images of the person bounding boxes. While earlier methods employed cross-entropy~\cite{Xiao2016LearningDF,Zheng2016MARSAV}, more recent and better performing ones adopt triplet loss~\cite{DBLP:journals/corr/HermansBL17,Luo_2019_CVPR_Workshops,Martinel2019a,Auto2019Ruijie} to learn to generate unique feature embeddings for each person ID. We apply CIR and improve performance of a plain triplet loss siamese-net approach~\cite{DBLP:journals/corr/HermansBL17} as well as of the current best method~\cite{Chen_2019_ICCV}.

\noindent \textbf{Person Search.}
This stands for the joint detection and re-identification of individuals in galleries of full images, given a single query~\cite{Xu2014PSS}. Most recent and best approaches use the Online Instance Matching (OIM)~\cite{xiao2017joint} to build up look tables of people ID representative embeddings~\cite{munjal2019cvpr,munjal2019bmvc,Chen_2018_ECCV,Xiao2017IANTI,Yan_2019_CVPR}. We apply CIR and slightly improve performance of the state-of-the-art work of \cite{munjal2019bmvc}, which employs the cross-entropy loss.

\begin{figure}[t!]
\centering
\includegraphics[trim=3cm 7.5cm 3cm 2.6cm, clip=true, width=0.8\linewidth]{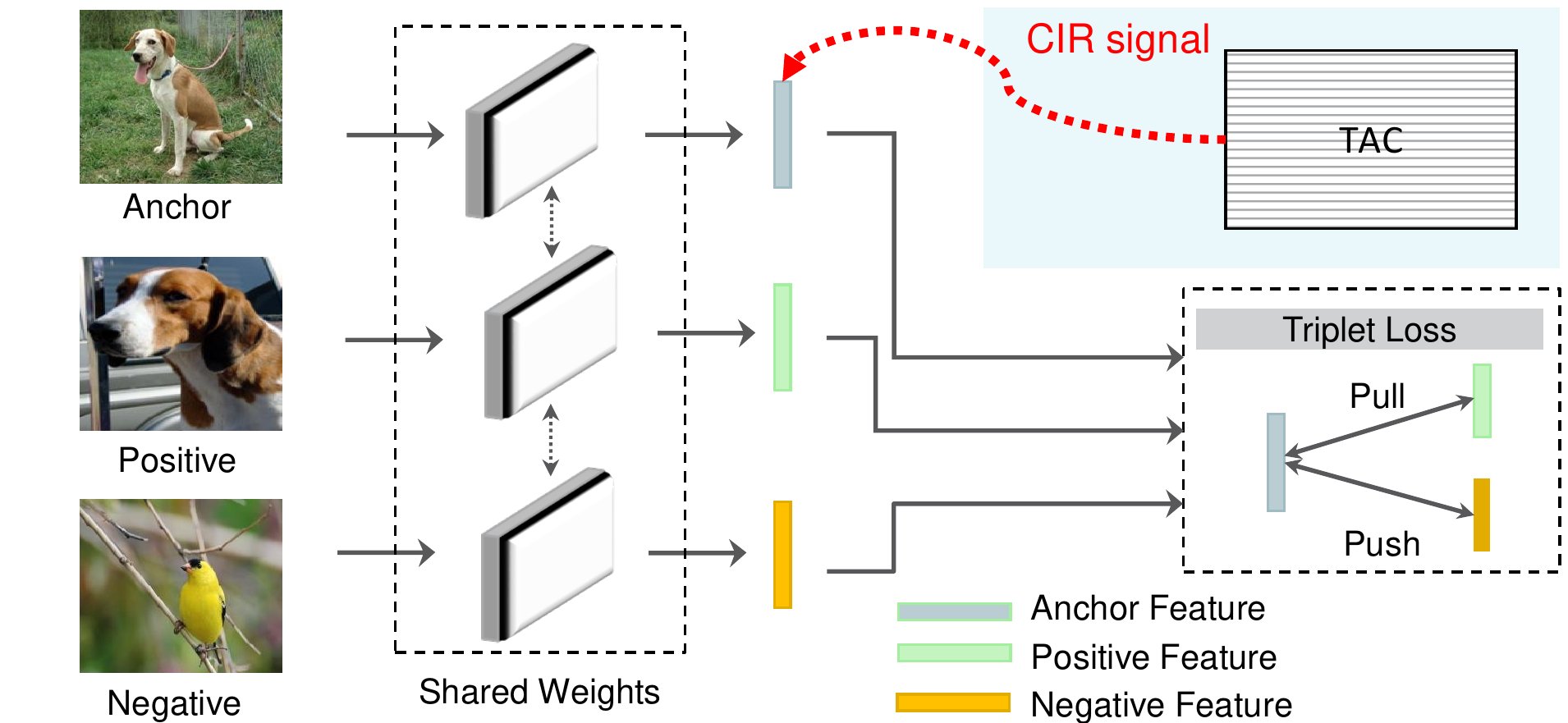}
\caption{\small Illustration of the proposed Class Interference Regularization (CIR), as applied to a triplet-Siamese neural network model with triplet loss.
During training, the average per-class output features are accumulated into a Table of Average Class embeddings (TAC) and used to perturb the anchor with the average embedding of a randomly-sampled wrong class.
The illustration refers to few-shot learning. Cropped person IDs are the input images in the case of person re-identification. The Cyan box shows the proposed CIR signal.
}
\label{fig:fewshot}
\end{figure}
\section{Class Interference Regularization (CIR)}
We propose CIR to regularize the training of multi-class DNNs, both for cross-entropy and for contrastive losses. In this section, we first introduce the class interference signal (cf.\ Fig.~\ref{fig:fewshot}); then we detail the implementation of CIR for the tasks of re-identification, few-shot learning and person search; finally we discuss CIR more formally and the intuition behind it.

\subsection{Class Interference}
CIR introduces a table $\Gamma \in \mathbb{R}^{C\mathrm{x}d}$ of average class embeddings into DNN models to track the mean embeddings $\mathbf{\mu}_c \in \mathbb{R}^d$ for each of the $C$ classes.
Then for each image $\textbf{x}_i$ with feature embedding $\mathbf{z}_i \in \mathbb{R}^d$ and class $y_i$, CIR randomly selects a mean class embedding $\mathbf{\mu}_c$ of another class $c\in C$ from $\Gamma$, with $c \neq y_i$, to corrupt $\mathbf{z}_i$. This yields a  new blended embedding
\begin{equation}
    \tilde{\mathbf{z}}_i = (1 - \lambda) \mathbf{z}_i + \lambda \mathbf{\mu}_c
    \label{eq:interference}
\end{equation}
where $\lambda \in [0, 1]$ controls the amount of interference.
In other words, the polluted embedding $\tilde{\mathbf{z}}_i$ is given by $\mathbf{z}_i$ ``pushed'' towards the mean embedding $\mathbf{\mu}_c$ of a wrong class $c$. This interference makes the optimization tougher and reduces overfitting. We further discuss CIR and provide an intuition to it in Sec.~\ref{sec:discussion}.

\begin{algorithm}[H]
\small
\caption{Application of CIR to person re-identification and few-shot learning with triplet loss.}
\textbf{Input:} \textit{Training data:} $\mathcal{D} = \{ (\textbf{x}_n, y_n) \}^N_{n=1}$, where $\textbf{x}_n$ represents an image, $y_n$ its corresponding person ID (re-id) or object class (few-shot learning);
\textit{hyper-parameters:} $\Gamma$ (TAC) update momentum $\gamma$, interference amount $\lambda$, margin $\delta$, max. iterations $T$, learning rate $\alpha$

\textbf{Initialization:} Triplet-siamese network model with initial parameters $\theta^{(0)}$ of the feature extractor $f(\textbf{x}_n, \theta^{(0)}) \in \mathbb{R}^d$, and $\Gamma^{(0)} \in \mathcal{R}^{C\mathrm{x}d}$ is randomly initialized TAC

\For{$t=1, \ldots , T$} {

$\mathcal{D}_t = \{ (\mathbf{a}_i, \mathbf{p}_i, \mathbf{n}_i)\}^B_{i=1} \gets$ {\small select a mini-batch of triplets of size $B$ from the training set}

where $\mathbf{a}_i, \mathbf{p}_i$ are from the same id/class $\mathbf{y}_i$ and $\mathbf{n}_i$ is from another identity

\For{$i=1, \ldots , B$} {
$\mathbf{z}_i^a = f(\mathbf{a}_i, \theta^{(t-1)}) \gets$ {\small compute feature embeddings for anchor}

$\mathbf{z}_i^p = f(\mathbf{p}_i, \theta^{(t-1)}) \gets$ {\small compute feature embeddings for positive}

$\mathbf{z}_i^n = f(\mathbf{n}_i, \theta^{(t-1)}) \gets$ {\small compute feature embeddings for negative}

$\mathbf{\mu}_c$ = $\Gamma^{(t-1)}[c] \gets${\small average embedding of randomly chosen class $c \neq y_i$ from TAC}
 
$\tilde{\mathbf{z}}_i^a = (1 - \lambda) \mathbf{z}_i^a + \lambda \mathbf{\mu}_c \gets$ {\small class interference acc. to Eq.~\ref{eq:interference} only for anchor embedding}

$\mathcal{L}_i = \mathcal{L}_{triplet}(\mathbf{z}_i^a, \mathbf{z}_i^p, \mathbf{z}_i^n, \delta) \gets$ {\small triplet loss acc. to Eq.~\ref{eq:triplet}}
}

$\theta^{(t)} \gets \theta^{(t-1)} + \alpha \frac{1}{\|\mathcal{D}_t\|} \sum_{i \in \mathcal{D}_t} \nabla_{\theta^{(t-1)}}(\mathcal{L}_i)$

$\Gamma^{(t)} = (1-\gamma) \Gamma^{(t-1)} + \gamma\   \textbf{z}_{\mathcal{D}_t} \gets$ {\small update TAC per class using corresponding embeddings from the current mini-batch during backward pass}
}
\textbf{Output:} Trained model parameters $\theta^{(T)}$
\label{alg:personreid}
\end{algorithm}

\subsection{Application to selected tasks}
We employ CIR for re-identification, few-shot learning and person search. 
In all cases, CIR is applied with minor changes and provides consistent performance improvements (cf.\ Sec.~\ref{sec:exp}).

\myparagraph{\textbf{Person Re-identification:}}
In re-identification (re-id) the model is tasked with the identification of the persons in the query, provided as crops.
State-of-the-art approaches in re-id employ the triplet loss for feature learning, which is formulated as:
\begin{equation}
    \mathcal{L}_{triplet}(\textbf{a}_i, \textbf{p}_i, \textbf{n}_j, \delta) = \max (0, \delta + \| \textbf{a}_i - \textbf{p}_i \|^2 - \| \textbf{a}_i - \textbf{n}_j \|^2 )
\label{eq:triplet}
\end{equation}
where $\textbf{a}_i$ and $\textbf{p}_i$ are anchor and positive samples, respectively, for the positive class $i$ and $\textbf{n}_j$ is the sample of the negative class $j$. 
While $\delta$ represents the expected margin between inter-class and intra-class distances.
For the task of re-id, we propose class interference as follows: first, we introduce a Table of Average Class embeddings (TAC) to accumulate the mean identity specific features; then, we add noise to the anchor $\textbf{a}_i$ according to Eq.~\ref{eq:interference} with randomly sampled mean class embedding from TAC. We outline the procedure in Algorithm~\ref{alg:personreid}.

\myparagraph{\textbf{Few-Shot Learning:}}
One-shot learning is in essence quite similar to re-identification as it aims to learn a model which is able to classify images having only seen one example per class.
For training, Prototypical loss is common in this case, or similar to re-identification Triplet-loss is also applicable.
In our initial experiments, we found that the performance of Proto lags behind the Triplet, therefore, we opt for the latter for our experiments. The use of Triplet loss also makes the application of CIR, in this case, similar to few-shot learning. Hence, the same algorithm\ref{alg:personreid} applies.

\myparagraph{\textbf{Person Search:}}
 OIM~\cite{xiao2017joint} is one of the most common approach for person search. Many recent state-of-the-art papers \cite{Liu2017NPSM,munjal2019bmvc,munjal2019cvpr} rely on OIM loss for feature learning.
During OIM training, the output feature of a person identity is matched against the TAC lookup table.
Using Eq.~\ref{eq:interference}, we corrupt the output features of the person identity with a randomly chosen mean embedding from the TAC. We outline the procedure in Algorithm~\ref{alg:personsearch}.

\begin{algorithm}[H]
\small
\caption{Application of CIR to person search.}
\textbf{Input:}
\textit{Training data:} $\mathcal{D} = \{ (\textbf{x}_n, \textbf{y}_n) \}^N_{n=1}$, where $\textbf{x}_n$ represents an image, bold-face $\textbf{y}_n$ is corresponding ground-truth person ID and the bounding box, while $y_n$ represents only the ground-truth person ID; 
\textit{hyper-parameters:} $\Gamma$ (TAC) update momentum $\gamma$, interference amount $\lambda$, max. iterations $T$, and learning rate $\alpha$

\textbf{Initialization:} OIM~\cite{xiao2017joint} network model with initial parameters $\theta^{(0)}$ of the feature extractor $f(\textbf{x}_n, \theta^{(0)}) \in \mathbb{R}^d$, and $\Gamma^{(0)} \in \mathcal{R}^{C\mathrm{x}d}$ is randomly initialized TAC

\For{$t=1, \ldots , T$} {

$\mathcal{D}_t = \{ (\mathbf{x}_i, \mathbf{y}_i)\}^B_{i=1} \gets$ {\small select a mini-batch of size $B$ from the training set}

\For{$i=1, \ldots , B$} {
$\mathbf{z}_i = f(\mathbf{x}_i, \theta^{(t-1)}) \gets$ {\small compute feature embeddings}

$\mathbf{\mu}_j$ = $\Gamma^{(t-1)}[c] \gets${\small average embedding of randomly chosen class $c \neq y_i$ from TAC}

$\tilde{\mathbf{z}}_i = (1 - \lambda) \mathbf{z}_i + \lambda \mathbf{\mu}_j \gets$ {\small class interference acc. to Eq.~\ref{eq:interference}}

$\hat{y}_i = \underset{c}{\arg\max} \Big(\Gamma^{(t-1)} \tilde{\mathbf{z}}_i \Big) \gets$ {\small predicted person ID}

$\mathcal{L}_i = \mathcal{L}_{ \text{FRCNN}}(\textbf{y}_i, \hat{\textbf{y}}_i) + \mathcal{L}_{\text{OIM}}(y_i, \hat{y}_i) \gets$ {\small loss for person search as in \cite{xiao2017joint}}
}
$\theta^{(t)} \gets \theta^{(t-1)} + \alpha \frac{1}{\|\mathcal{D}_t\|} \sum_{i \in \mathcal{D}_t} \nabla_{\theta^{(t-1)}}(\mathcal{L}_i)$

$\Gamma^{(t)} = (1-\gamma) \Gamma^{(t-1)} + \gamma\  \textbf{z}_{\mathcal{D}_t} \gets$ {\small update TAC per class using corresponding embeddings from the current mini-batch during backward pass}
}
\textbf{Output:} Trained model parameters $\theta^{(T)}$
\label{alg:personsearch}
\end{algorithm}

\subsection{Intuition and discussion on CIR}\label{sec:discussion}

We explain the regularizing effect of CIR on the network training by a simple study case. Assume the regression task of learning an image embedding $\mathbf{z}_i$ for the $i$-th image $\mathbf{x}_i$, according to the target ground-truth embedding $\mathbf{y}_i$. We would assume that $\mathbf{z}_i$ be the result of a simple linear relation, i.e.\ 1-layer fully-connected network, $\mathbf{z}_i = W \mathbf{x}_i$\,. In the equation, $W$ are the current network parameters.
The regression loss is given by
\begin{equation}
    \displaystyle
    \mathcal{L}(W,\mathbf{x}_i,\mathbf{y}_i) =
    \frac{1}{2} \left \| W \mathbf{x}_i - \mathbf{y}_i \right \| ^2 =
    \frac{1}{2} \left \| \mathbf{z}_i - \mathbf{y}_i \right \| ^2
\end{equation}

When applying CIR, we substitute for $\mathbf{z}_i$ with $(1 - \lambda) \mathbf{z}_i + \lambda \mathbf{\mu}_c$, as provided by Eq.~\eqref{eq:interference}.
This yields a regularized loss $\displaystyle\mathcal{L}_{\text{CIR}}$ given by:
\begin{equation}
    \displaystyle
    \mathcal{L_{\text{CIR}}}(W,\mathbf{x}_i,\mathbf{y}_i) =
    \frac{1}{2} \left \| (1 - \lambda) \mathbf{z}_i + \lambda \mathbf{\mu}_c - \mathbf{y}_i \right \| ^2 =
    \frac{1}{2} \left \| (\mathbf{z}_i - \mathbf{y}_i) -\lambda (\mathbf{z}_i  -  \mathbf{\mu}_c  ) \right \| ^2
\label{eq:cir}
\end{equation}

Notice that by allowing interference we actually force $\mathbf{z}_i$ to come closer to the average embedding $\mathbf{\mu}_c$ of the wrong class. However, the optimization as a result tries to push the classes even further apart so that even a noisy embedding stays far away from the average embedding of other classes.
\camr{We further support this intuition with an analysis of feature embeddings in Sec.~\ref{sec:expfewshot}}

\vspace{-0.5cm}
\section{Experiments}
\label{sec:exp}

Here we first evaluate CIR on algorithms based on triplet losses for Few Shot Learning and Person Re-identification; then we benchmark it on algorithms for Person Search and Classification adopting cross-entropy losses.

\subsection{Few Shot Learning}\label{sec:expfewshot}
\myparagraph{Dataset and metrics.} We consider the \textbf{miniImageNet}~\cite{oriol2016nips} and \textbf{tieredImageNet}~\cite{ren18fewshotssl} datasets. The first is a subset of ILSVRC-12 dataset~\cite{ILSVRC15} with 100 classes in total and 600 images per class. miniImageNet is divided into 64, 16, and 20 classes for meta-training, meta-validation, and meta-testing, respectively. 
tieredImageNet~\cite{ren18fewshotssl} is a larger and more complex subset of ILSVRC-12 with hierarchical structure. It contains 608 classes and 779,165 images in total. The classes are grouped into 34 broader categories according to WordNet~\cite{imagenet_cvpr09} with 20 training, 6 validation, and 8 testing subsets. Following \cite{snell2017nips}, we report the classification accuracy by taking the average over 600 randomly generated episodes from test set.\\
\myparagraph{Implementation Details.} Our work is based upon an open-source implementation of the ProtoNet\footnote{\label{footnote:web} ProtoNet implementation for few shot learning\\ \url{https://github.com/wyharveychen/CloserLookFewShot}} with Resnet18~\cite{he2016resnet} as backbone architecture. However, we use a triplet-siamese network with triplet loss for optimization. We train our model in two settings, from scratch (one-stage) and from a pre-trained model using softmax cross-entropy loss over all classes (two-stage). Note that for pre-training we do not use any extra data from the original ImageNet dataset. For triplet from scratch, we train for 300 epochs and decay the learning rate exponentially~\cite{DBLP:journals/corr/HermansBL17} after 200 epochs. For triplet pre-trained, we train for 100 epochs and decay the learning rate after 50 epochs.
Each epoch has 100 iterations and the initial learning rate is set to 0.0002. We use online triplet mining with \textit{Batch All} sampling strategy as suggested in~\cite{DBLP:journals/corr/HermansBL17}.  For miniImagenet, we use a batch size of 80 (20 classes, 4 samples per class) and for tieredImagenet we use batch size 256 (64 classes, 4 samples per class). We use update momentum $\gamma$=0.5 for building the TAC and regularization momentum $\lambda$=0.5 for class interference. \camr{The value of TAC update momentum is motivated from \cite{xiao2017joint}}.
Note that, apart from keeping a TAC for average embeddings of different classes during training, there are no additional computational and memory space overheads, considering corruption of signal is negligible.\\
\myparagraph{Results.} 
In Table~\ref{tab:mini_tiered_reid}(a), we show the results of our evaluations in comparison to the state-of-the-art.
We prefer triplet over prototypical loss for our baseline due to its superior performance (cf. Table~\ref{tab:mini_tiered_reid}(a)). We first evaluate the triplet-siamese network in one stage setting (training from scratch). On miniImageNet dataset, this model achieves an accuracy of 57.4\% for 1 shot and 70.2\% for 5 shot. For tieredImagenet, this model achieves 63.9\% for 1 shot and 77.1\% for 5 shot. As shown in the table, the addition of CIR to this model, improves its performance on miniImagenet by approximately 1pp (58.4 vs 57.4) for 1 shot and 2.4pp (72.6 vs 70.2) for 5 shot. For tieredImagenet, we observe marginal improvement. 

We then evaluate our model in two-stage setting (cross-entropy followed by triplet) and report results in the last section of the table. As shown, this model provides a very strong baseline and CIR further improves this strong baseline significantly. For miniImagenet, CIR brings an improvement of almost 2.7pp for 1 shot and 1.5pp for 5 shot. For tieredImagenet, CIR brings an improvement of 2.5pp for 1 shot and 1.2pp for 5 shot. In the same Table~\ref{tab:mini_tiered_reid}(a), we also list the results of the state-of-the art models on few-shot learning. On tieredImagenet we outperform the current best approaches LEO~\cite{rusu2018metalearning} by 2.8pp on 1-shot learning and MetaOptNet~\cite{lee2019meta} by 1.4pp on 5-shot learning.

\begin{table}[t]
\small
\begin{center}

\begin{tabular}{cc}
\resizebox{7cm}{!}{
\begin{tabular}{lcccc}

& \multicolumn{2}{c}{{miniImageNet}} & \multicolumn{2}{c}{{tieredImageNet}}\\
\hline
Method & \textbf{1-shot}  & \textbf{5-shot} & \textbf{1-shot} & \textbf{5-shot}  \\
\hline
\hline
ProtoNet~\cite{snell2017nips}, NIPS17 & 49.4  & 68.2 & 53.3 & 72.7\\
ProtoNet~\cite{chen2019closerlook} (ResNet18), ICLR18 &  54.2 &  73.7 & -& -\\
Large Margin (Triplet) \cite{wang2018large}, NIPS18 & 50.1 &  66.9& - & -\\

Relation Net \cite{sung2018cvpr}, CVPR18 & 50.4 &  65.3& 54.5&71.3 \\
Transductive Prop \cite{DBLP:journals/corr/abs-1805-10002}, ICLR19 & 55.5 & 69.9& 59.9  & 73.3 \\
Incremental \cite{DBLP:journals/corr/abs-1810-07218}, NIPS19 & 54.9  & 63.0  & 56.1 & 65.5\\
IDeMe-Net~\cite{chen2019deform}, CVPR19   & 57.7 & 74.3 & - & -\\
Individual Feature~\cite{gordon2018metalearning}, ICLR19 & 56.9 & 70.5 & -& - \\
CAML \cite{jiang2018learning}, ICLR19 & 59.2 & 72.3  & - & - \\
LEO \cite{rusu2018metalearning}, ICLR19 & 61.8  & 77.6  & \textbf{66.3} & 81.4\\
MetaOptNet \cite{lee2019meta}, CVPR19 & \textbf{62.6}  & \textbf{78.6} & 65.9 & \textbf{81.5}\\
Global Class Rep. \cite{li2019few}, ICCV19 &  53.2 & 72.3 & - & -\\

\hline
Triplet&  57.4 & 70.2 & 63.9 & 77.1 \\
+ \textit{CIR} & \textbf{58.4} & \textbf{72.6} & \textbf{64.3} & \textbf{77.7}\\

 \hline
Cross Entropy $\rightarrow$ Triplet & 57.7 & 74.3 & 66.6 & 81.7\\
 + \textit{CIR} & \textbf{60.5} & \textbf{75.8} & \textbf{69.1} & \textbf{82.9} \\
\hline
\end{tabular}

}
&

\resizebox{5.3cm}{!}{
\begin{tabular}{lcc}
& \multicolumn{2}{c}{{Market-1501}} \\
\hline
 Method & \textbf{mAP(\%)}  & \textbf{Rank-1(\%)}  \\
\hline
\hline
\dag Triplet~ \cite{DBLP:journals/corr/HermansBL17}, arXiv17 &  69.1 & 84.9 \\
\dag MGN~\cite{2018arXiv180401438W}, ACM18 &   86.9 & 95.7   \\
\dag SSP-ReID~\cite{quispe2019enhanced}, IMAVIS19 &{75.9} & {89.3} \\

HPM~\cite{fu2018horizontal}, AAAI19 & 82.7 &94.2 \\
\dag PyrNet~\cite{Martinel2019a}, CVPR19 & 81.7 & 93.6  \\
\dag BoT~ \cite{Luo_2019_CVPR_Workshops}, CVPR19 & 85.9 & 94.5    \\
DGNet~\cite{zheng2019joint}, CVPR19  & 86.0 & 94.8  \\
CASN~\cite{meng2019reid}, CVPR19  & 82.8 & 94.4  \\
\dag Auto-ReID~\cite{Auto2019Ruijie}, ICCV19 &  85.1 & 94.5 \\
\dag OSNet~\cite{zhou2019osnet}, ICCV19 & 84.9 & 94.8 \\
\dag P$^{2}$-Net~\cite{Guo_2019_ICCV}, ICCV19 & 85.6 & 95.2 \\
\hline

Triplet ours & 70.4 & 84.5 \\

Triplet ours + \textit{CIR}& \textbf{71.3} & \textbf{85.8}\\

\hline
\dag ABD-Net~\cite{Chen_2019_ICCV}, ICCV19&  88.3 & \textbf{95.6}  \\
\dag ABD-Net~\cite{Chen_2019_ICCV} + \textit{CIR}&   \textbf{88.8} & \textbf{95.6} \\	

\hline
\end{tabular} 
}\\
\small{(a)}& \small{\ \ (b)} \\
\end{tabular}
\vspace{-0.6cm}

\end{center}

\caption{\small (a) CIR on few shot learning for 5-way 1-shot and 5-way 5-shot. The average accuracy of 600 randomly generated episodes is reported for miniImageNet~\cite{oriol2016nips} and tieredImageNet~\cite{ren18fewshotssl} (b) Comparative evaluation of CIR for the task of person re-identification on the Market-1501 dataset~\cite{zheng2015scalable}. Methods indicated as \dag\ also use triplet loss and could also benefit from the simple CIR strategy.}
\label{tab:mini_tiered_reid}
\end{table}

\camr{Furthermore, we perform a sanity check by adding gaussian noise to the feature embeddings instead of CIR, as a regularizer. 
These results are shown in Table~\ref{tab:noise} on tieredImageNet for 5-way 1-shot case. 
We notice that adding gaussian noise does not have any impact on the results, whereas CIR shows consistent improvements in all cases. 
This experiment allows us to understand that the contribution of CIR is more significant than just the random noise.
}

\begin{figure}[t]
\begin{center}
    
\begin{tabular}{cc}
\includegraphics[width=60mm,height=35mm]{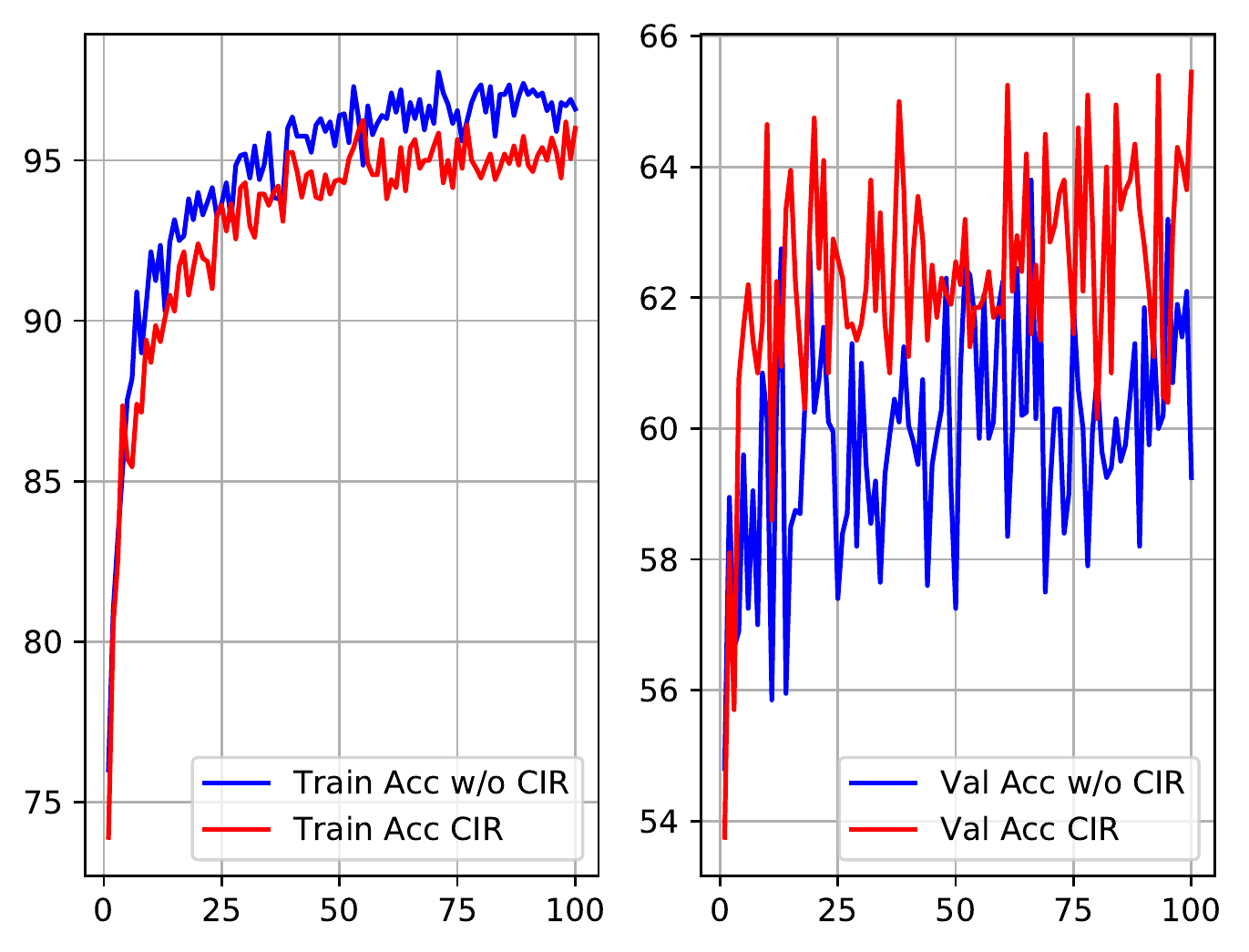} &   \includegraphics[width=60mm,height=35mm]{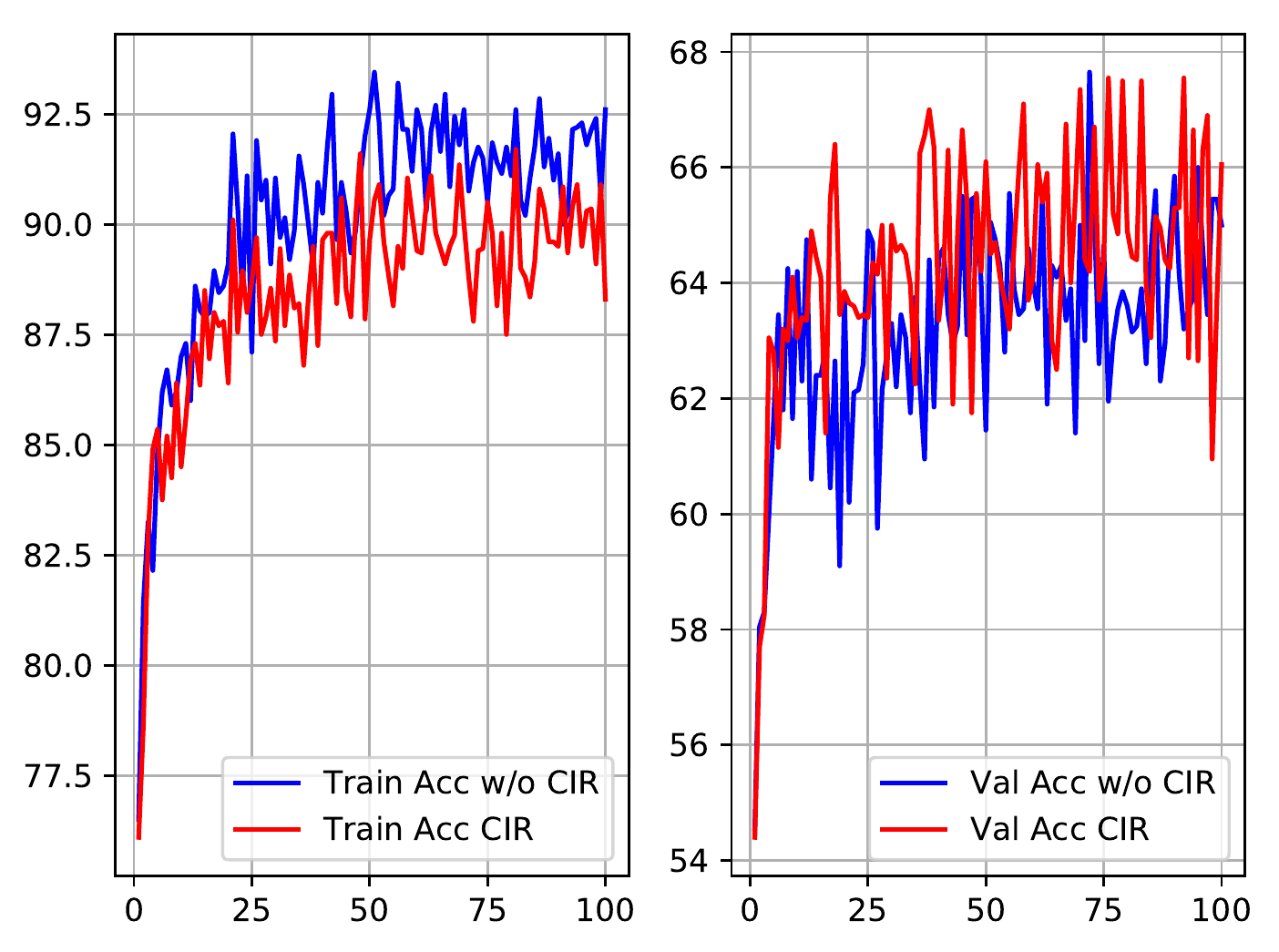} \\
\small{\ \ (a)}& \small{\ \ (b)} \\
\end{tabular}
\vspace{-0.5cm}

\end{center}

\caption{\small Training and validation accuracies with or without CIR on miniImagenet (a), tieredImagenet (b). Notice, improvement in validation accuracy and slight drop in training accuracy with CIR, indicating reduction in overfitting.}
\label{fig:overfiitting_curves}
\end{figure}

\myparagraph{CIR as a Regularizer}: We empirically demonstrate in Figure~\ref{fig:overfiitting_curves} that CIR acts as a regularizer. Figure~\ref{fig:overfiitting_curves}(a) shows the plots for training (left) and validation (right) accuracy on miniImagenet. Similarly, Figure~\ref{fig:overfiitting_curves}(b) shows the plots for training (left) and validation (right) accuracy on tieredImagenet. As shown for both the datasets, our proposed CIR model has constantly higher validation accuracy than the one without CIR, while having lower training accuracy. This shows that CIR has a regularizing effect as it is able to generalize beyond the training set and decrease the gap in training and validation accuracy.

\camr{\myparagraph{Effect of CIR on the feature embedding}:
To better support the intuition on CIR of Sec.~\ref{sec:discussion}, with reference to tieredImagenet~\cite{ren18fewshotssl}, we compute the average distance from the overall center of mass of each data-point embedding and the ratio of inter-to-intra class mean distances. The first increases from 6.99 (w/o CIR) to 38.88 (w/ CIR), while the second changes from 1.02 (w/o CIR) to 1.17 (w/ CIR). This means that CIR effectively makes the feature-embedding space expand, but the embeddings from each class remain relatively compact.}

\begin{table}[t]
\footnotesize
\begin{center}
\begin{tabular}{lc}

\hline
Method & tieredImageNet (1-shot)  \\
\hline
\hline

\hline
Triplet & 63.9 \\
Triplet + \textit{Gaussian Noise} & 63.6 \\
Triplet + \textit{CIR} & \textbf{64.3}\\

 \hline
Cross Entropy $\rightarrow$ Triplet & 66.6 \\
Cross Entropy $\rightarrow$ Triplet + \textit{Gaussian Noise} & 66.9 \\
Cross Entropy $\rightarrow$ Triplet + \textit{CIR} & \textbf{69.1} \\
\hline
\end{tabular}
\end{center}

\caption{\small Few-shot learning results for CIR \emph{vs} Gaussian Noise on tieredImageNet [52]. The numbers represent the average accuracy of 600 randomly generated episodes.}
\label{tab:noise}
\end{table}

\subsection{Person Re-identification}
\myparagraph{Dataset and metrics.} We adopt the \textbf{Market-1501}~\cite{zheng2015scalable} dataset, which contains a total of 32,668 images representing the cropped bounding boxes of 1,501 persons. The train/test splits contain 750 and 751 identities respectively. For evaluation, we use the standard metrics, mAP and CMC rank-1.\\
\myparagraph{Implementation Details.} Triplet is the most common loss used in person re-id literature due to its superior performance. 
We re-implement our triplet baseline for person re-identification following \cite{DBLP:journals/corr/HermansBL17}. We use pre-trained ResNet-50 architecture with input images re-scaled to $256 \times 128$. For augmentation, random crops and horizontal flipping are applied during training. For CIR, we employ TAC update momentum $\gamma$=0.5 \camr{as discussed in Section~\ref{sec:expfewshot}} and regularization momentum $\lambda$=0.1.\\
\myparagraph{Results.}
In Table~\ref{tab:mini_tiered_reid}(b), we show the results of our baseline re-implementation ``Triplet ours'' are slightly better than the original work~\cite{DBLP:journals/corr/HermansBL17}. We show that our proposed regularization CIR improves the baseline by 0.9pp mAP and 1.3pp CMC Rank-1.
Note that other techniques are also available for person re-identification with better performance than~\cite{DBLP:journals/corr/HermansBL17}, however most of them are still based on triplet loss. We add CIR on top of ABD-Net~\cite{Chen_2019_ICCV} which is the state-of-the-art in person re-identification and also uses triplet loss. 
As shown in the table, our proposed CIR brings an improvement of 0.5pp mAP and provides the best known mAP score for person re-identification on the Market-1501 dataset.

\begin{table}[t]
\small
\begin{center}

\begin{tabular}{cc}
\resizebox{7cm}{!}{
\begin{tabular}{lcc}
& \multicolumn{2}{c}{{CUHK-SYSU}} \\
\hline
Method & \textbf{mAP(\%)} & \textbf{top-1(\%)} \\
\hline
\hline
OIM~\cite{xiao2017joint}, CVPR17  & 75.5 & 78.7 \\
IAN~\cite{Xiao2017IANTI}, arXiv17  & 76.3 & 80.1  \\
NPSM~ \cite{Liu2017NPSM}, ICCV17  & 77.9 & 81.2 \\
\dag Mask-G~\cite{Chen_2018_ECCV}, ECCV18  & 83.0 & 83.7\\
CLSA~\cite{person_eccv18}, ECCV18 & 87.2 & 88.5\\

\dag QEEPS~\cite{munjal2019cvpr}, CVPR19 &{84.4} & {84.4} \\
\dag Context Graph~\cite{Yan_2019_CVPR}, CVPR19 & 84.1 & 86.5\\
\hline
OIM ours  & 77.8 & 78.1 \\
OIM ours + \textit{CIR} & \textbf{79.3} & \textbf{80.0}  \\
\hline

\dag Distilled QEEPS~\cite{munjal2019bmvc}  (Resnet18), BMVC19 & 84.1 & 84.3\\
\dag Distilled QEEPS~\cite{munjal2019bmvc} (Resnet18) +  \textit{CIR} & \textbf{84.5} & \textbf{84.6}\\

\hline

\hline

\end{tabular}

}
&

\resizebox{5.3cm}{!}{
\begin{tabular}{cccc}

\hline

{CIFAR-10} & Accuracy & Accuracy-test \\
\hline
No Reg.~\cite{rafael2019nips}& $86.8\pm0.2$ &  86.8 \\
Label Smoothing~\cite{rafael2019nips}& 86.7$\pm$0.3 & 87.0\\
CIR  & - & 87.1\\
\hline
{CIFAR-100}  & Accuracy & Accuracy-test \\
\hline
No Reg.~\cite{rafael2019nips}& $72.1\pm0.3$ & 75.2 & \\
Label Smoothing~\cite{rafael2019nips}& 72.7$\pm$0.3 & 76.0 & \\ 
CIR & - & 76.1\\
\hline

\end{tabular}
}\\
\small{(a)}& \small{\ \ (b)} \\
\end{tabular}
\vspace{-0.5cm}

\end{center}

\caption{\small (a) Comparative evaluation of {CIR} for the task of person search on the CUHK-SYSU [59] dataset. CIR boosts performance of OIM by a significant margin by regularizing its training. Methods indicated as \dag\ are built on top of OIM and could also benefit from the simple CIR strategy. (b) Top-1 Accuracy on CIFAR-10 and CIFAR-100 datasets. We follow the same implementation as in~\cite{rafael2019nips}. Accuracy-test is the accuracy of our implementation on actual test set.}
\label{fig:cuhk_cifar}
\end{table}

\subsection{Person Search}
Most recent approaches for person search are based on OIM~\cite{xiao2017joint} model. Hence, we also consider this as our baseline approach.
To apply CIR in this case, we blend the ID feature embedding of a person with the average embedding from some other person ID from TAC.\\
\myparagraph{Dataset and metrics.} We adopt most commonly used \textbf{CUHK-SYSU}~\cite{xiao2017joint} dataset for benchmarking with 18,184 images labeled, 8,432 identities and 96,143 bounding boxes. We adopt the train/test split of \cite{xiao2017joint}. The dataset presents challenging large variations in person appearance, background clutter and illumination changes. As metrics, we follow \cite{xiao2017joint} and adopt mean Average Precision (mAP) and Common Matching Characteristic (CMC top-1).\\
\myparagraph{Implementation Details.} We re-implement the OIM person search algorithm of~\cite{xiao2017joint} in Pytorch, which we consider as our baseline. We use an image resolution of 600 pixels (shorter side). For CIR, we employ TAC update momentum $\gamma$=0.5 \camr{as discussed in Section~\ref{sec:expfewshot}} and regularization momentum $\lambda$=0.5.\\
\myparagraph{Results.}
We report in Table~\ref{fig:cuhk_cifar}(a) the most recent relevant results together with ours. Our baseline ``OIM ours'' shows 77.8 mAP, slightly above the original OIM performance~\cite{xiao2017joint}. Implementing the CIR regularization on top of it yields 79.3 mAP and 80.0 top-1 CMC, improving the metrics by 1.5pp and 1.9pp, respectively. As shown in the table, most of these approaches~\cite{Chen_2018_ECCV,Yan_2019_CVPR,munjal2019cvpr,munjal2019bmvc} are based on OIM, therefore CIR is directly applicable to them. We add CIR to the state-of-the-art person search method of Distilled QEEPS~\cite{munjal2019bmvc}, which uses OIM. We adopt ResNet18 as the backbone of the Distilled QEEPS and use hyper-parameters of the original paper~\cite{munjal2019bmvc}.
As shown in the Table~\ref{fig:cuhk_cifar}(a), the proposed CIR brings an improvement of 0.4pp mAP and 0.3pp top-1 on Distilled QEEPS~\cite{munjal2019bmvc}.

\subsection{Classification}
Finally we demonstrate the application of CIR for the classification models trained with traditional cross-entropy loss. This also allows us to compare to label smoothing regularization~\cite{rafael2019nips}.

\noindent \textbf{Dataset and metrics.} We consider \textbf{CIFAR-10} and \textbf{CIFAR-100} which are widely adopted datasets for natural image recognition. Both datasets are subsets from 80-million tiny image database~\cite{Torralba0880million} and contain 60k images ($32 \times 32$) each. The train split has 50k images and test set has 10k images. CIFAR-10 has 10 categories while CIFAR-100 has 100.

\noindent \textbf{Implementation Details.} We use AlexNet~\cite{NIPS2012_4824} for CIFAR-10 and ResNet-56~\cite{he2016resnet} for CIFAR-100, as suggested in~\cite{rafael2019nips}. However, for a transparent evaluation and reproduciblity of results by the community, we evaluate our model on publicly available test set (Accuracy-test in Table~\ref{fig:cuhk_cifar} (b)), unlike~\cite{rafael2019nips} which defines its own private validation set for both CIFAR-10 and CIFAR-100. For both datasets, we employ CIR with TAC update momentum $\gamma$=0.5 \camr{as discussed in Section~\ref{sec:expfewshot}} and regularization momentum $\lambda$=0.1.

\noindent \textbf{Results.} 
In Table~\ref{fig:cuhk_cifar}, we notice that both Label Smoothing~\cite{christian2016cvpr} and our proposed CIR give minor but consistent improvement for both CIFAR-10 and -100, over model without regularization. In future work, we aim to study further applications for our proposed CIR.

\vspace{-0.5cm}
\section{Conclusions}
\label{sec:conclusions}

We have proposed CIR, a novel, simple and effective regularization technique. CIR applies to cross entropy losses but is especially suited to contrastive losses. CIR is the first to act on the output features and it parallels established regularization techniques acting on the label distributions such as label smoothing and self-distillation, which do not apply to contrastive losses. In experimental evaluation we have shown that CIR improves consistently the performance of few-shot learning and person re-identification -- for contrastive losses -- and person search and classification -- for cross-entropy losses. In the latter case, improvements are more modest, but on par with label smoothing. Given the rising popularity of contrastive losses and given the simplicity of CIR, we hope that it would play a role in future model trainings.


\bibliography{egbib}
\end{document}